\documentclass{article} 
\usepackage{PRIMEarxiv}
\usepackage{url}  
\usepackage{graphicx}
\usepackage{caption} 

\usepackage{algorithm}
\usepackage{algorithmic}

\usepackage{wrapfig}

\usepackage{newfloat}
\usepackage{color}
\usepackage{amsmath}
\usepackage{todonotes}

\newcommand{\D}{\mathcal{D}}
\newcommand{\X}{\mathcal{X}}
\newcommand{\Y}{\mathcal{Y}}

\newcommand{\test}{\text{test}}
\newcommand{\train}{\text{train}}

\title{The Role of Active Learning in Modern Machine Learning}
\author{
	Thorben Werner 
	\thanks{Institute of Computer Science - Information Systems and Machine Learning Lab (ISMLL)} 
	\\
	University of Hildesheim\\
	Universitätsplatz 1\, 31141 Hildesheim \\
	\texttt{werner@ismll.de} \\
	\And
	Prof. Lars Schmidt-Thieme$^*$ \\
	University of Hildesheim\\
	Universitätsplatz 1, 31141 Hildesheim \\
	\texttt{schmidt-thieme@ismll.uni-hildesheim.de}
	\And
	Dr. Vijaya Krishna Yalavarthi$^*$ \\
	University of Hildesheim\\
	Universitätsplatz 1, 31141 Hildesheim \\
	\texttt{yalavarthi@ismll.uni-hildesheim.de}
}

\begin{document}

\maketitle

\begin{abstract}
Even though Active Learning (AL) is widely studied, it is rarely applied in contexts outside its
own scientific literature. 
We posit that the reason for this is AL's high computational cost coupled with the comparatively small lifts 
it is typically able to generate in scenarios with few labeled points.
In this work we study the impact of different methods to combat this low data scenario, namely data augmentation (DA), 
semi-supervised learning (SSL) and AL.
We find that AL is by far the least efficient method of solving the low data problem, generating a lift of only 1-4\% 
over random sampling, while DA and SSL methods can generate up to 60\% lift in combination with random sampling.
However, when AL is combined with strong DA and SSL techniques, it surprisingly is still able to provide improvements.
Based on these results, we frame AL not as a method to combat missing labels, but as the final building block to 
squeeze the last bits of performance out of data after appropriate DA and SSL methods as been applied.
\end{abstract}

\section{Introduction}
Training ML models in most real use cases entails working with limited amounts of labeled data.
Since labels are expensive to obtain, datasets usually are split into a small labeled pool and
a much larger unlabeled pool. \\
In this paper we provide insights about the three most researched techniques to train strong models 
under these constraints: data augmentation (DA), semi-supervised learning (SSL) and active learning (AL).
Even though all three techniques work differently (DA increases the amount of labeled data, SSL makes use
of unlabeled data and AL tries to improve the selection of points that are labeled), all of them solve the 
same problem of limited availability of labeled data.
In this sense, AL directly competes with DA and SSL as strategies for enhancing model quality in low-label regimes.
Current literature has yet provided a comprehensive study of the combined application of all three methods,
researching the question of which method works best in isolation, as well as whether they can be freely 
combined (with each consecutive method still providing a lift). 
In this work, we employ two well known DA methods and a collection of well-performing AL algorithms from a 
recent benchmark \cite{werner2024cross}.
As SSL paradigm, we chose pretraining as the most used SSL paradigm in recent literature. \\
We pay special attention to the performance of active learning methods, as techniques like DA or SSL are
very rarely used in AL literature. 
\begin{figure}[h]
	\centering
	\includegraphics[width=0.33\linewidth]{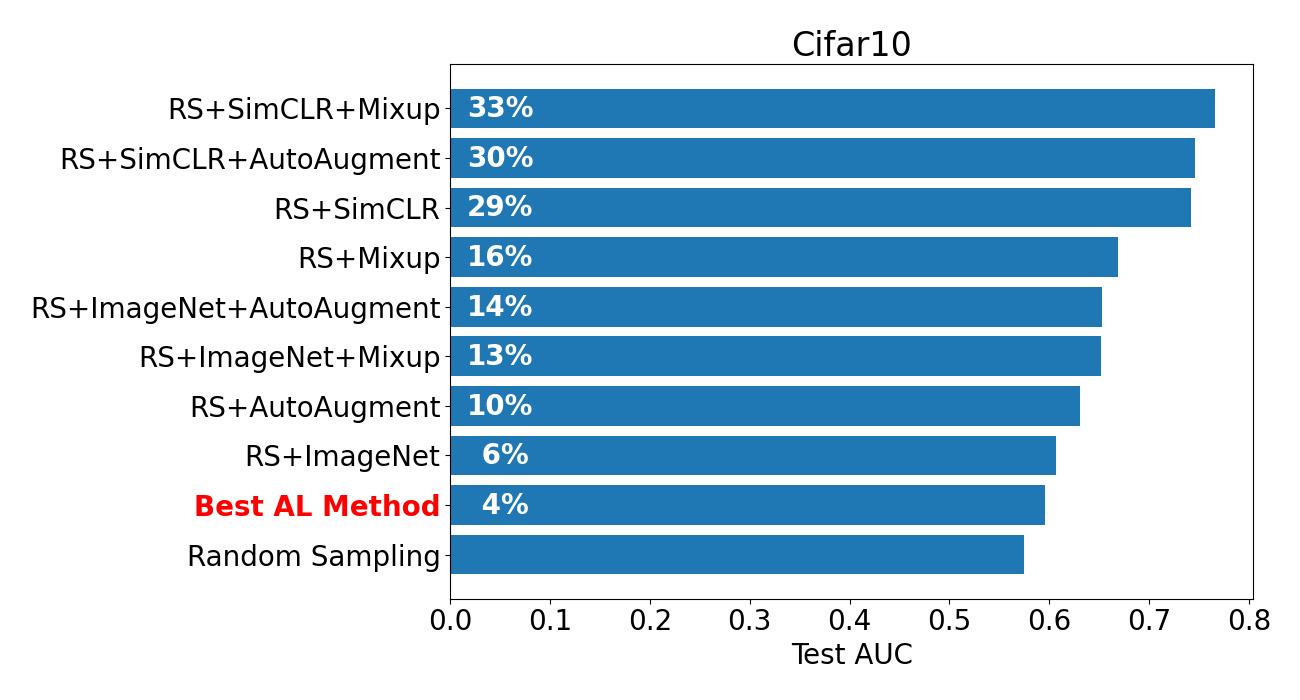}
	\includegraphics[width=0.33\linewidth]{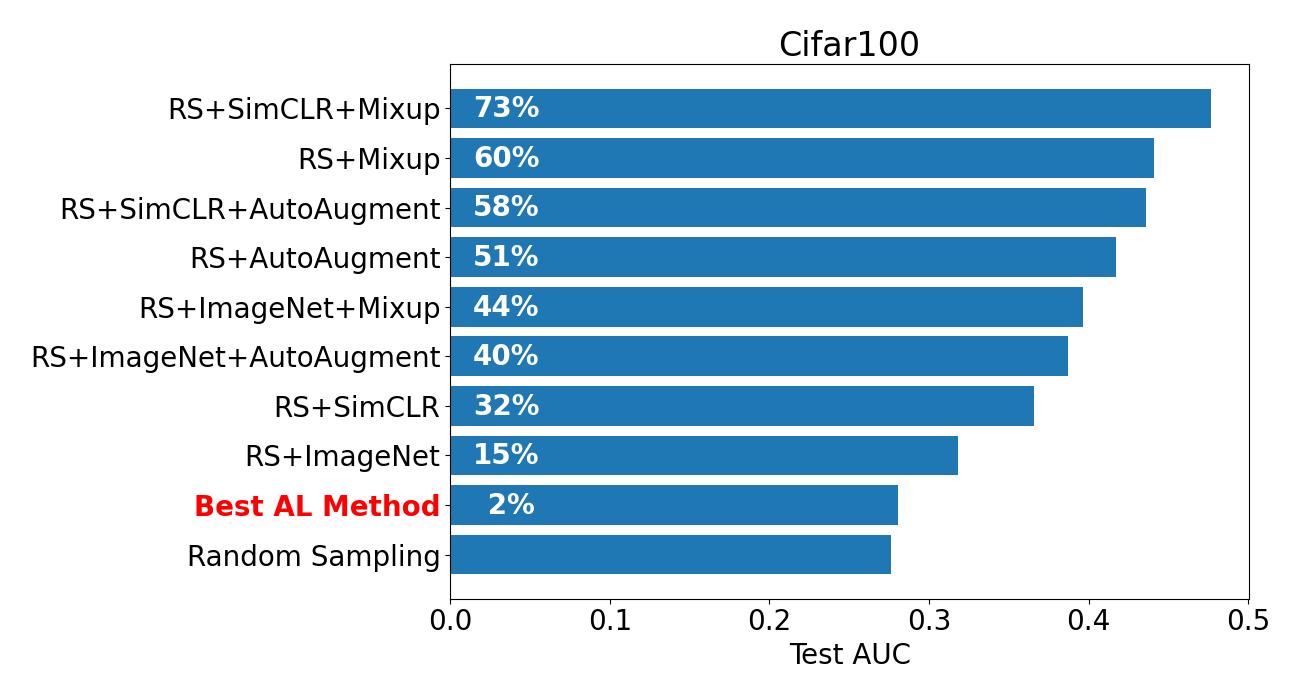}
	\includegraphics[width=0.33\linewidth]{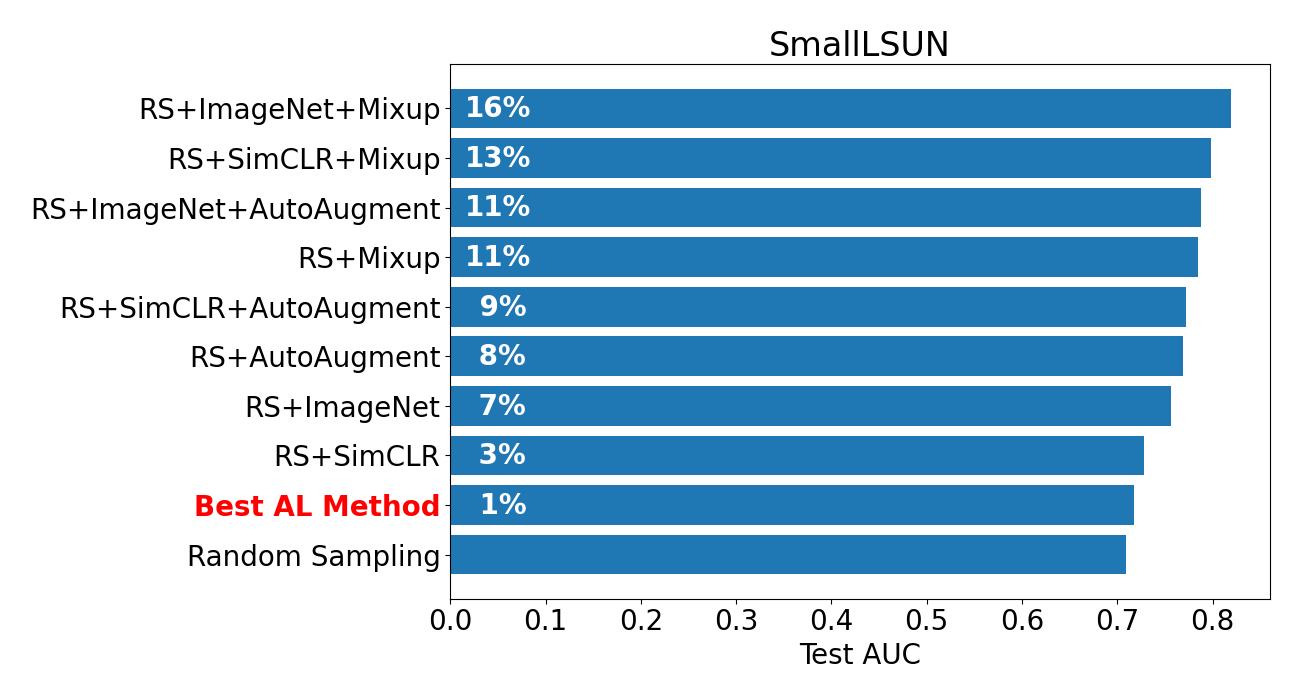}
	\caption{Performance of Random Sampling (RS) plus different DA and SSL methods and the best AL method without
	DA or SSL. Numbers in each bar indicate the percentage improvement over random sampling.}
	\label{fig:advantage_plots}
\end{figure}
As a motivating example, we are comparing random sampling with various advanced training protocols containing 
DA and/or SSL against the \textbf{best} performing AL algorithm without DA or SSL on that dataset.
From Fig. \ref{fig:advantage_plots} you can observe that active learning techniques fall behind DA 
or SSL methods in terms of how much lift they provide over a randomly sampled labeled set and plain supervised 
learning without augmentations. \\
This elicits the question, whether AL is a useful technique to combat low data scenarios at all, or if DA and SSL
already exhaust the available lifts.
To the best of our knowledge, no methodological paper in AL has tested their proposed algorithm in a regime with 
strong DA and SSL methods and only one benchmark paper \cite{luth2024navigating} provides tertiary experiments
about the combination of all three methods.
Considering the high computational cost of AL, a systematic study on the impact of AL on a modern training pipeline
including DA and SSL provides a valuable answer to the question "Do we need AL in modern machine learning?".
To this end, we study different combinations of DA and SSL techniques for three different datasets
and test, whether AL methods can provide an additional lift on the optimal setup per dataset. 
\footnote{Code will be available under: TODO}
%
\subsection*{Key Insights}
\begin{enumerate}
	\item Active Learning is the least efficient method of overcoming low data scenarios
	\item Despite that, active learning can still provide lifts even when paired with strong data augmentation 
	and semi-supervised learning techniques
	\item In this regime, every competitive active learning method performs exactly on par
\end{enumerate}

\section{Problem Description}\label{sec:problem_description}
We are experimenting on pool-based AL with classification models.
Mathematically we have the following: \\
Given a dataset $\D_\train := (x_i, y_i) \quad i \in \{1, \ldots, N\}$ with $x \in \X, y \in \Y$ (following
\cite{werner2024cross} we similarly have 
$\D_\text{val}$ and $\D_\text{test}$) we randomly sample an initial labeled pool $L^{(0)} \sim \D_\text{train}$ that we call the seed set.
We suppress the labels from the remaining samples to form the initial unlabeled pool $U^{(0)} = \D_\text{train} / L^{(0)}$.
We define an acquisition function to be a function that selects a batch of samples of size $\tau$ from the unlabeled pool 
$a(U^{(i)}) := \{x_b^{(i)}\} \in U^{(i)} \quad b := [0, \ldots, \tau]$.
We then recover the corresponding labels $y^{(i)}_b$ for these samples and add them to the labeled pool 
$L^{(i+1)} := L^{(i)} \cup \{(x_b^{(i)}, y^{(i)}_b)\}$ and $U^{(i+1)} := U^{(i)} / \{x_b^{(i)}\} \quad b := [0, \ldots, \tau]$.
The acquisition function is applied until a budget $B$ is exhausted. \\ [1mm]
We measure the performance of a model $\hat{y}: \X \to \Y$ on the held out test set $\D_\text{test}$ after each 
acquisition round by fitting the model $\hat{y}^{(i)}$ on $L^{(i)}$ and measuring the test accuracy. \\
We allow the fitting process to additionally depend on a DA technique and an SSL technique who aid model training: 
$\text{TRAIN}(\hat{y}^{(i)}, \\L^{(i)}, \text{DA}, \text{SSL})$.
DA is allowed to alter the labeled samples of that iteration: $\text{DA}(L^{(i)})$, while SSL can make use of either 
a fully labeled external dataset $\D_\text{ext} \cap \D_\train = \emptyset$ and/or the unlabeled pool of our current dataset:
$\text{SSL}(\hat{y}^{(i)}, \{\D_\text{ext}, U^{(i)}\})$.

\section{Related Work}
We are currently not aware of any methodological paper that combines strong DA and SSL with their proposed method, as they usually
focus on improving upon other acquisition functions in a comparable setting.
However, some recent benchmark papers have studied aspects of DA and SSL:
\cite{beck2021effective} found that DA does not only improve the overall test accuracy of BADGE, but also its label efficiency.
\cite{werner2024cross} test AL methods in an SSL setting by pre-encoding their datasets with a pretrained encoder, but do not
employ DA in any of their experiments.
\cite{luth2024navigating} propose to tune DA as part of the hyperparameters in the first iteration of AL, as well as 
evaluating AL in two SSL scenarios.
We extend the study of \cite{luth2024navigating} in three ways: First, by allowing a comprehensive evaluation of DA and SSL techniques
on the tested datasets in order to find the optimal combination, second, by quantifying how much each method contributes
to overcoming the low data problem, and third, by significantly extending the list of tested AL algorithms.

\section{Methodology}\label{sec:methodology}
This work serves as a guide for machine learning practitioners tasked with training high-performing models 
on unlabeled datasets. In many cases, random sampling and simple techniques like data augmentation (DA) and leveraging 
pretrained ImageNet weights are the only methods applied, due to their low computational cost and accessibility.
We study the impact of various DA and semi-supervised learning (SSL) techniques when used 
alongside random data selection, and explore whether active learning (AL) can provide additional improvements in 
these settings. \\ [1mm]
We argue that the effectiveness of an AL method is not necessarily independent of the presence of DA or SSL. 
Marginal improvements from AL (as seen in Fig. \ref{fig:advantage_plots}) may be overshadowed by stronger techniques.
To address this, we propose a series of three experiments: (i) Analyzing the individual ability of DA, SSL and AL 
of improving upon the random sampling baseline with vanilla supervised training, (ii) finding the optimal combination
of DA and SSL for each dataset and (iii) testing whether AL is able to provide a lift over random sampling combined with
optimal DA and SSL. \\ [1mm]
To evaluate our experiments, we measure test accuracy of our classifier in $i$ rounds, where each classifier $\hat{y}^{(i)}$
is trained on $L^{(i)}$ with $i \in [1 \ldots B/\tau]$.
Each round we add 500 samples to our labeled pool ($\tau = 500$).
As aggregate metric we are using the normalized area under the accuracy curve (AUC):
\begin{equation}
	\text{AUC}(\D_\test, \hat y, B) := \frac{1}{B/\tau} \sum\limits_{i=1}^{B/\tau} \text{Acc}(\D_\test, \hat y^{(i)})
\end{equation}
A higher AUC signifies better average performance across $i$ rounds of testing. 
Note that this protocol is also followed for experiments using only random sampling in combination with DA or SSL.
Even though we could randomly sample $B$ points all at once and train a single model, we opt for a unified protocol
and the use of AUC values for two reasons:
(i) AUC is the preferred method of evaluating iterative AL algorithms like BADGE and, this way, we obtain directly
comparable results and 
(ii) the AUC is less dependent on the chosen budget.
A comparison based on the final accuracy for any high budget might be meaningless for lower budgets of practical applications.
The AUC incorporates this information in its score.
Furthermore, we repeat every experiment 20 times and compare the results with paired-t-tests and Critical Difference diagrams,
adhering to the best practices proposed by \cite{werner2024cross}.
Additionally, we report the learning trajectories of all tested methods in App. \ref{app:remaining_results}. 
The investigated methods are AutoAugment \cite{cubuk2019autoaugment} and Mixup \cite{zhang2017mixup} for DA and 
pretrained ImageNet weights and SimCLR \cite{chen2020simple} for SSL. 
For AL, we incorporate all well-performing AL methods from \cite{werner2024cross}, namely 
Badge \cite{ashdeep}, Galaxy \cite{zhang2022galaxy}, Uncertainty Sampling (Entropy, Margin, Least Confident), 
Coreset \cite{sener2017active} and CoreGCN \cite{caramalau2021sequential}.
For a summary of employed datasets and their chosen budgets, refer to Table \ref{tab:datasets}.
Please note, that we can not use any dataset that is derived from ImageNet, as we are using ImageNet-weights as a pretraining
method for our classifers.
\begin{table}
	\centering
	\begin{tabular}{l|c c c}
			& Cifar10 & Cifar100 & SmallLSUN \\
		\hline
		Query Size & 500 & 500 & 500 \\
		Budget & 10k & 30k & 20k \\
		\#Classes & 10 & 100 & 6 \\
		Imgs per Class & 6000 & 600 & $\sim$10k \\
		Img Size & 32 & 32 & 224
	\end{tabular}
	\caption{Statistics of the employed datasets. SmallLSUN is composed of 6 classes from the Large-scale Scene 
	Understanding (LSUN) dataset \cite{wang2017knowledge}. For details, please refer to Appendix \ref{app:lsun}.}
	\label{tab:datasets}
\end{table}

\newpage
\section{Implementation Details}\label{sec:impl_details}
\begin{wrapfigure}{r}{0.5\textwidth}
	\centering
	\includegraphics[width=\linewidth]{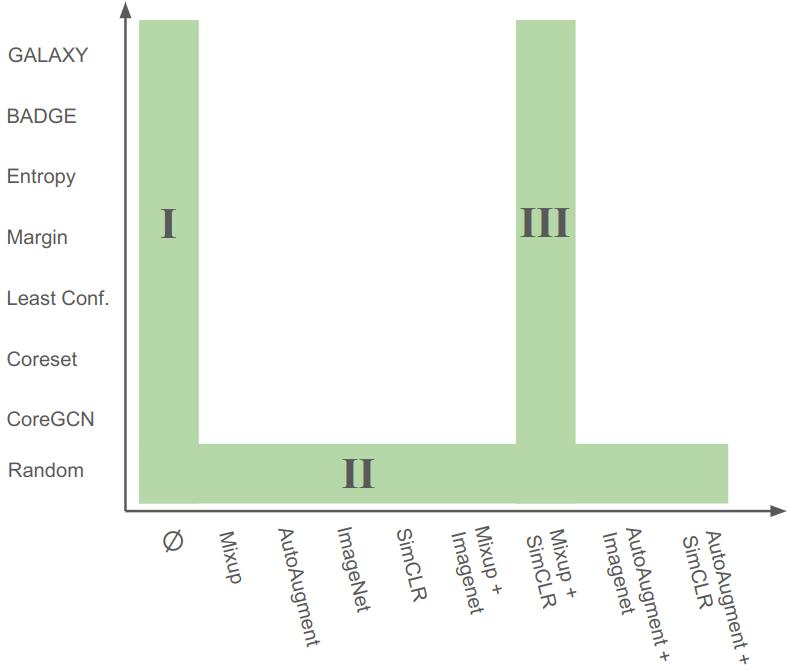}
	\caption{Overview of conducted experiments on Cifar100 with a ResNet18. Green areas indicate tested combinations.}
	\label{fig:experiment_grid}
\end{wrapfigure}In this work, we are using ResNet18 as our classification model to rule out any dirt effects from unstable training methods
or unexpected drops in performance for a novel dataset. 
We deliberately choose not to optimize our hyperparameters in a search, but rather use the default settings of our
chosen optimizer. \\
This is on one hand adhering to the validation paradox described in \cite{luth2024navigating}, where an optimal set of hyperparameters for AL 
cannot be found as this would entail labeling excessive amounts of extra data, on the other hand we argue that uncertainty
sampling methods profit unproportionally from optimized hyperparameters.
Evidence of this can be seen in the recent benchmark paper of \cite{werner2024cross}, where Least Confidence Sampling 
was the best performing AL method in the vision domain, and Margin Sampling being the best method over all. 
In order to enable a fair comparison between uncertainty- and diversity-based AL methods, we use default choices for our
hyperparameters.
For further details of our employed hyperparameters, refer to Appendix \ref{app:hyperparameters}. \\
Our SimCLR pretraining is identical to \cite{werner2024cross} with 100-200 epochs of unsupervised contrastive training with
SGD and a cosine learning rate scheduler.

\section{Experiments}\label{sec:experiments}

\begin{figure}[ht]
	\centering
	\includegraphics[width=0.8\linewidth]{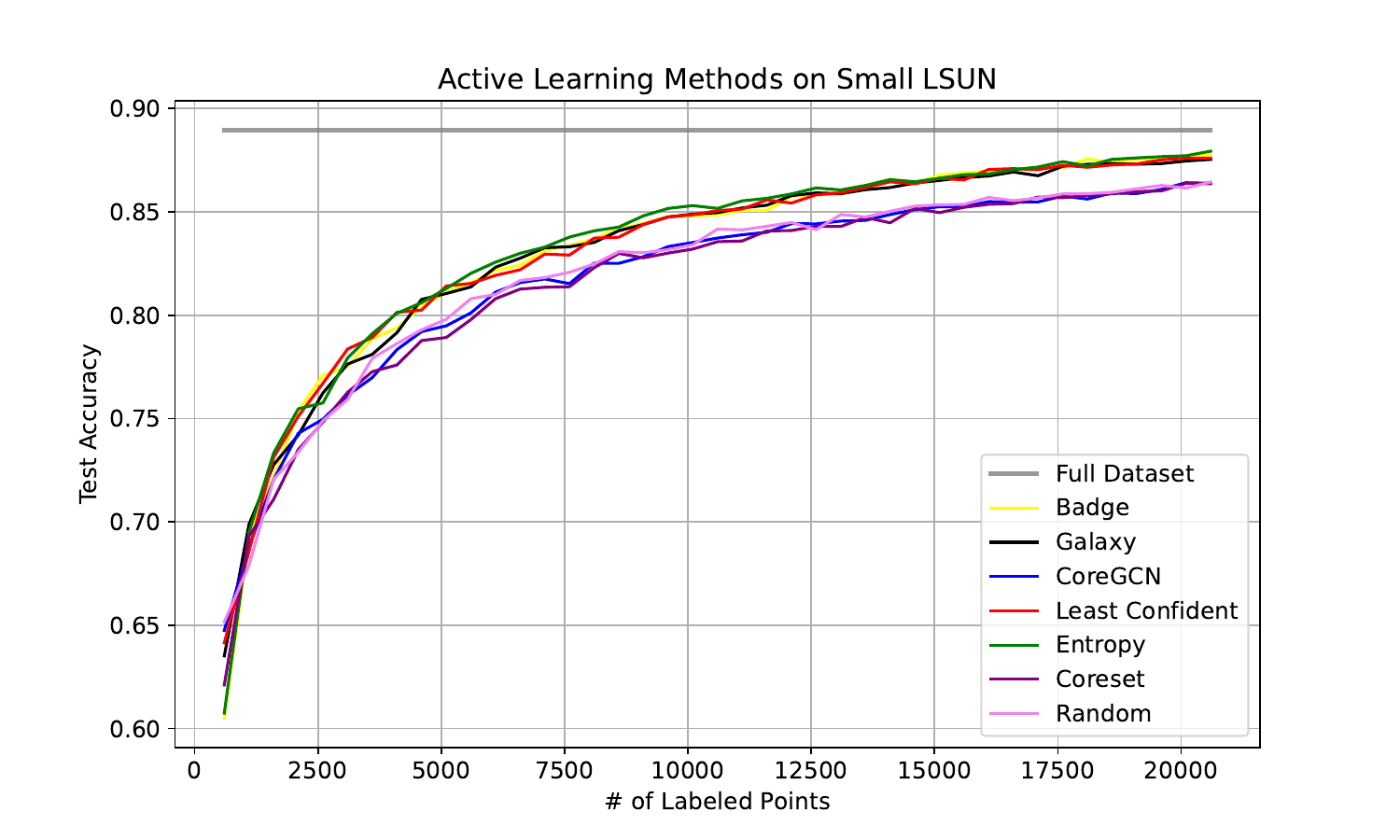}
	\includegraphics[width=0.7\linewidth]{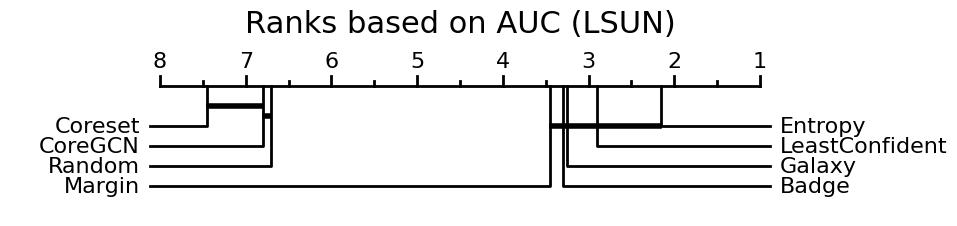}
	\caption{Test accuracy curves for all AL methods for LSUN on a setup with ImageNet and Mixup (top) and the resulting 
	Critical Difference diagram based in the AUC values (bottom)}
	\label{fig:results_lsun}
\end{figure}

\begin{figure}[ht]
	\centering
	\includegraphics[width=0.8\linewidth]{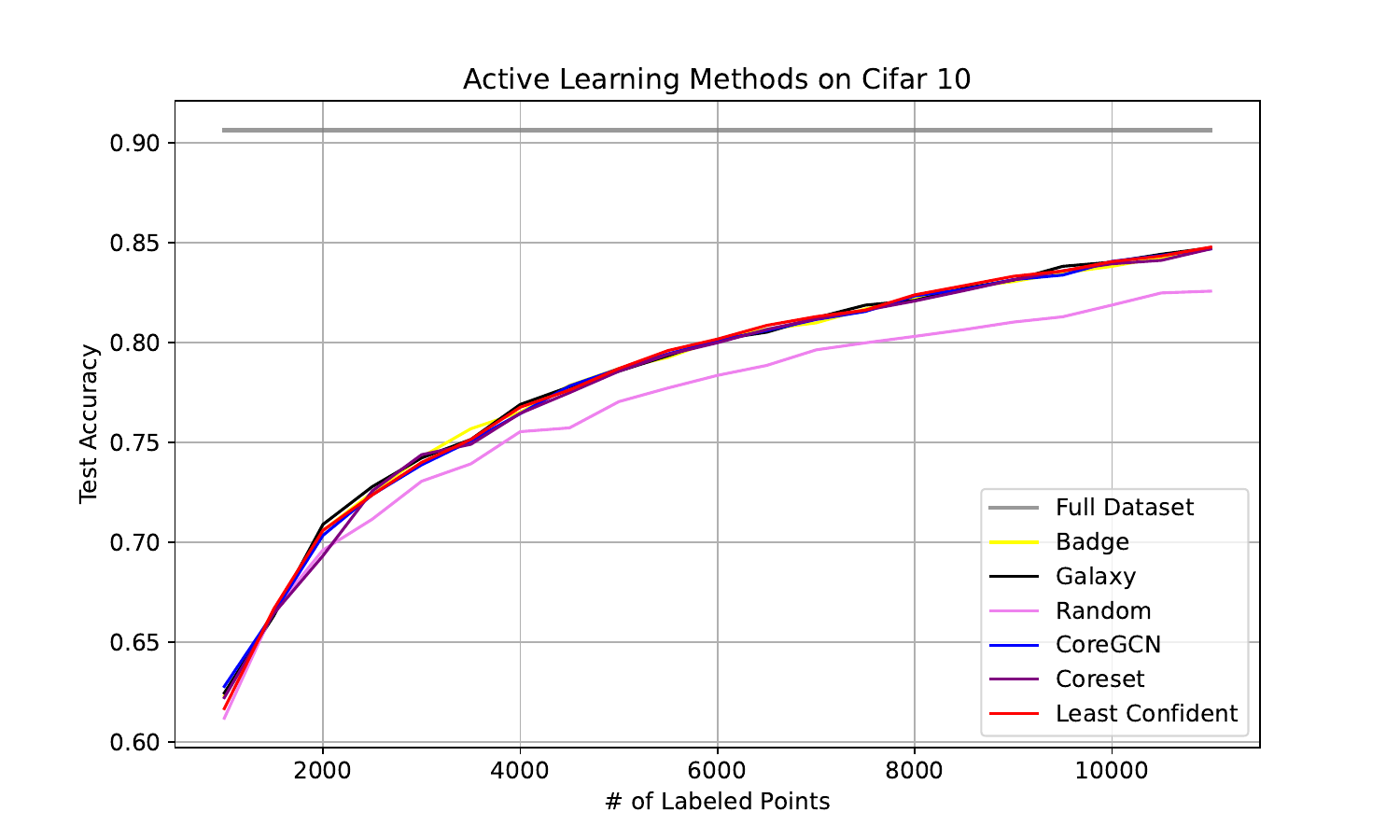}
	\includegraphics[width=0.7\linewidth]{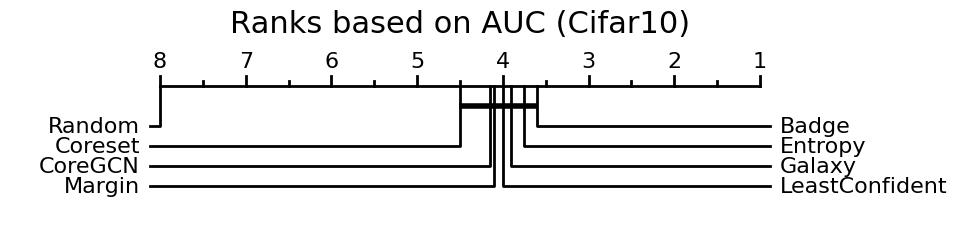}
	\caption{Test accuracy curves for all AL methods for Cifar10 on a setup with ImageNet and Mixup (top) and the resulting 
	Critical Difference diagram based in the AUC values (bottom)}
	\label{fig:results_cifar10}
\end{figure}

\begin{figure}[ht]
	\centering
	\includegraphics[width=0.8\linewidth]{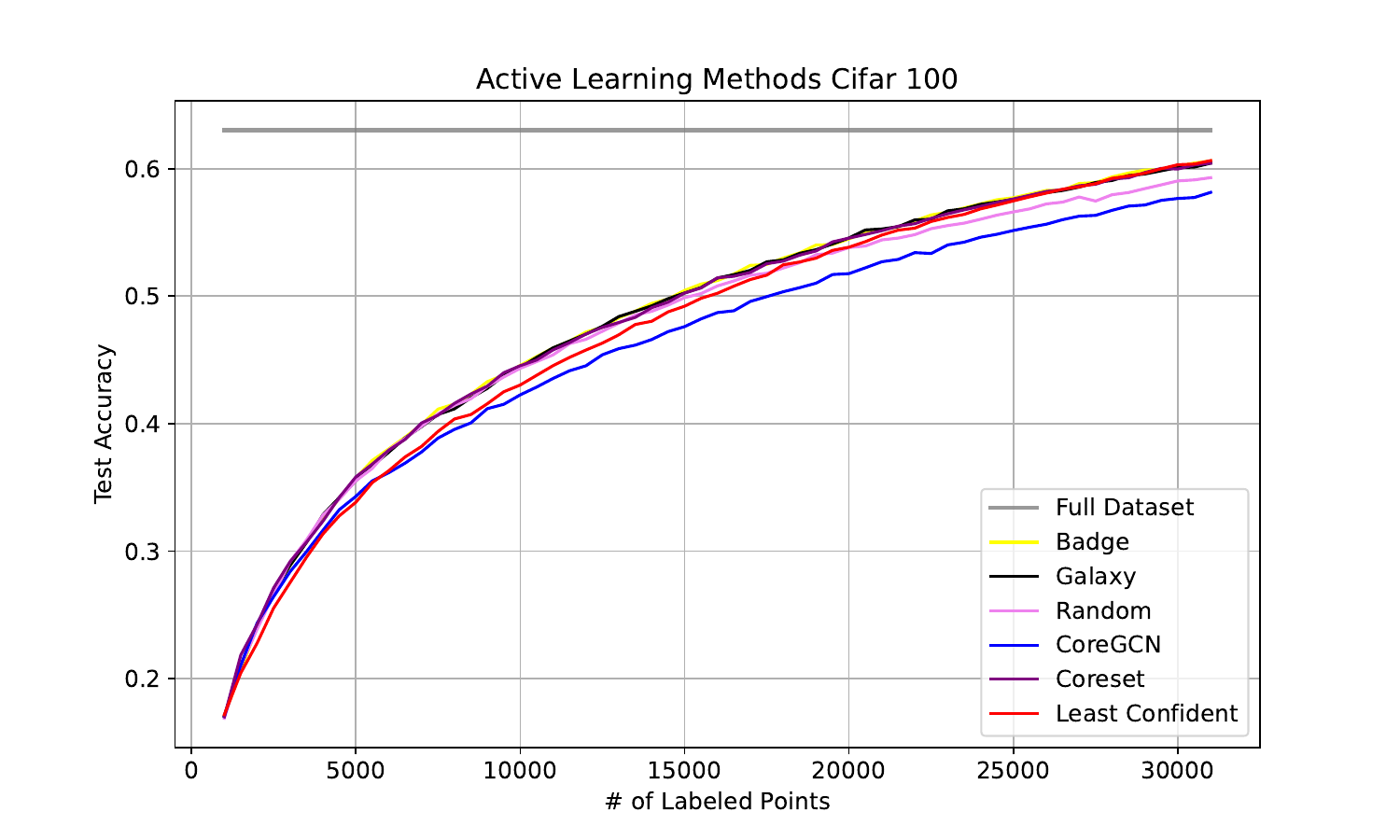}
	\includegraphics[width=0.7\linewidth]{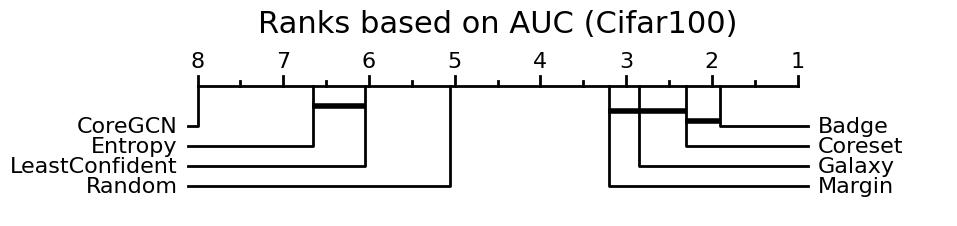}
	\caption{Test accuracy curves for all AL methods for Cifar100 on a setup with ImageNet and Mixup (top) and the resulting 
	Critical Difference diagram based in the AUC values (bottom)}
	\label{fig:results_cifar100}
\end{figure}

An overview of our conducted experiments can be found in Fig. \ref{fig:experiment_grid}. 
Areas shaded in green indicate tested combinations, while other regions have been omitted due to prohibitive computational costs.
First, we measure individual lifts of DA, SSL and AL methods (area I and II) by comparing their AUC values to the baseline AUC of random
sampling with vanilla supervised training.
\begin{figure}[ht]
	\centering
	\includegraphics[width=0.7\linewidth]{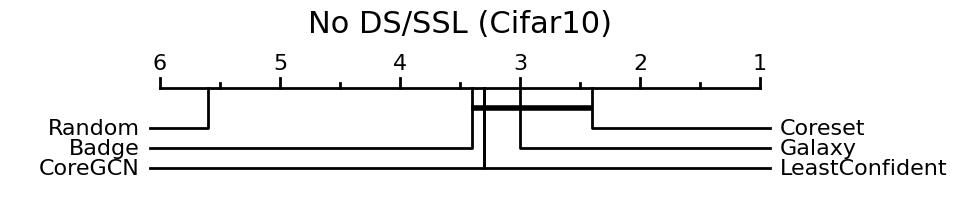}
	\includegraphics[width=0.7\linewidth]{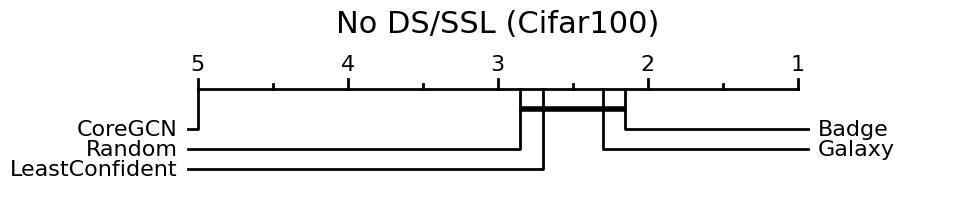}
	\includegraphics[width=0.7\linewidth]{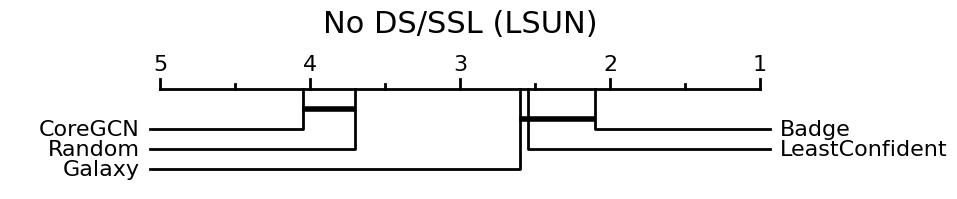}
	\caption{All tested AL methods on a training pipeline without DA or SSL (Area I
	in Fig. \ref{fig:experiment_grid})}
	\label{fig:vanilla_agents}
\end{figure}
We display the performance of all AL methods without DA or SSL in Fig \ref{fig:vanilla_agents} and 
the performance of each DA and SSL method in Fig. \ref{fig:advantage_plots}. 
The number in each bar indicates the percentage lift over the baseline.
We observe that AL methods significantly lag behind all other methods to combat low data scenarios.
This is especially impactful, considering the computational cost of the average AL setup.
Since AL requires training of $B/\tau$ many classifiers, it is roughly $B/\tau$ times more expensive than any DA technique or
ImageNet weights.
A direct comparison to the computational cost of SimCLR challenging, as the pretraining time varies between datasets.
In our case, the pretraining took longer than a single run of e.g. BADGE sampling, but this might change depending on 
chosen hyperparameters.
At the same time, SimCLR is offering a greater lift than any tested AL method.
From this experiment we conclude that AL alone is not efficient in overcoming the low data scenario, as much cheaper 
techniques with greater lift could always be employed. \\ [1mm]
Our second observation from Fig \ref{fig:advantage_plots} is that DA and SSL techniques generally stack well, i.e. they 
do not overshadow each other's lifts.
This is no novel insight, as many modern training pipelines for vision datasets successfully include both DA and SSL techniques.
AL literature, however, is rarely incorporating either, let alone both, eliciting the question, whether the lifts of 
AL methods also stack similarly.
To this end, we tested our AL methods on the best performing combination of DA and SSL for each dataset (area III in Fig. \ref{fig:experiment_grid}) 
creating the hardest possible environment to produce further lifts and display the results in 
Fig. \ref{fig:results_lsun}-\ref{fig:results_cifar100}.
From Fig. \ref{fig:results_lsun} we can clearly observe that some AL methods are still able to perform better than random sampling. 
Historically strong AL methods that rely on uncertainty sampling outperform random sampling with statistical 
significance, indicated by the missing bar between "Random" and "Margin" in Fig. \ref{fig:results_lsun} (bottom).
On the other hand, diversity-based methods like Coreset or CoreGCN do not consistently outperform random sampling,
although this behavior varies between datasets (Compare Fig. \ref{fig:results_cifar100}).
Finally, we observe that for the cluster of well-working methods per dataset, no method has any advantage above competing
methods.
We define "well-working" as being better than random sampling with a statistically significant lift.
Even though the critical difference diagrams indicate an advantage of Entropy Sampling in Fig \ref{fig:results_lsun} and 
Badge in Fig. \ref{fig:results_cifar10_100}, the difference in test accuracy is marginal (Fig. \ref{fig:results_lsun} (top)) 
Accuracy curves for Cifar10/100 are qualitatively the same and can be found in App \ref{app:remaining_results}.

\section{Conclusion}
In this work we quantified the individual impact of DA, SSL and AL methods on small randomly sampled, labeled pools.
We found that AL is by far the least efficient method of improving upon the low data scenario, since it offers only a 1-4\% 
lift over random sampling with a significant investment of compute.
A practitioner of machine learning can always employ appropriate DA and SSL techniques first and expect a better return. \\ [1mm]
After testing all AL methods with the best DA and SSL setup for each dataset, we found that while some methods improved 
performance, none had a significant best performance on any dataset.
The only consistent algorithms were Badge, Galaxy and Margin sampling; so after applying DA and SSL, a practitioner 
is free to choose among these. 
This is consistent with \cite{luth2024navigating}, who also found that the field of AL methods moves closer together 
in the presence of DA and SSL and some methods like Coreset start to collapse to random or sub-random performance. \\
This work serves as a guide for practitioners to design their training pipelines in a zero-shot manner:
While DA and SSL are universally beneficial and oftentimes cheap to obtain (even SimCLR only has to be done once), they
can decide to include AL based on the required performance on the dataset.
Only if they need to obtain the final few percentage points of the possible performance on this dataset, they 
should opt for AL. \\ [2mm]
We would like to close with a proposed paradigm shift in AL research:
Developing an AL method in an environment without DA and SSL techniques is not scientifically sound, as it might interact 
with these techniques in unpredictable ways.
Modern AL research needs to make sure that their proposed method does not collapse to random performance in modern training 
pipelines, and it should strive to outperform the cluster of strong uncertainty sampling methods that have been identified
by this work and recent benchmarks \cite{werner2024cross,luth2024navigating,ji2023randomness} in environments with DA and SSL. \\

\newpage

\bibliographystyle{plain}
\bibliography{main.bib} 

\appendix

\section{Small LSUN Data}\label{app:lsun}
SmallLSUN is composed of 6 classes from the Large-scale Scene Understanding (LSUN) dataset \cite{wang2017knowledge}.
The dataset contains $224 \times 224$ images of indoor and outdoor scenes with labels referring to the location of
the scene.
We have compiled a list of classes from this dataset that have more than 10k examples, but remain under 16GB of data
to limit the computational burden of testing AL algorithms. \\
We selected the following classes from \url{https://www.tensorflow.org/datasets/catalog/lsun}:
\begin{enumerate}
	\item Bridge
	\item Church outdoor
	\item Conference Room
	\item Dining Room
	\item Restaurant
	\item Tower
\end{enumerate}
Sampled 10k images from these 6 classes, resulting in a dataset of size 60k (train), 1.8k (validation) and 6k (test).
\begin{figure}[H]
	\centering
	\includegraphics[width=0.6\linewidth]{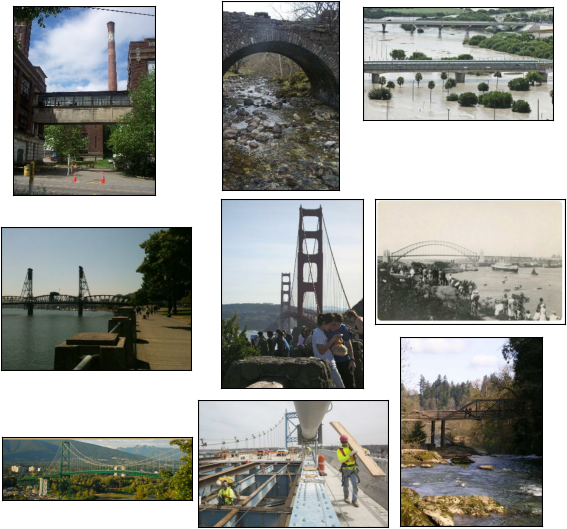}
	\caption{Example images for the Bridge class. Taken from {https://www.tensorflow.org/datasets/catalog/lsun}}
\end{figure}

\section{Hyperparameters}\label{app:hyperparameters}
\begin{table}[H]
	\centering
	\begin{tabular}{l | c c c}
			& Cifar10 & Cifar100 & LSUN \\
		\hline
		\hline
		\textbf{Evaluation} \\
		\hline
		Optimizer & NAdam & NAdam & NAdam \\
		Learning Rate & 0.001 & 0.001 & 0.001 \\
		Weight Decay & 0 & 0 & 0 \\
		\hline
		\hline
		\textbf{SimCLR} \\
		\hline
		Epochs & 100 & 100 & 250 \\
		Optimizer & SGD & SGD & SGD \\
		Initial LR & 0.4 & 0.4 & 0.4 \\
		LR Scheduler & Cosine & Cosine & Cosine \\
		Weight Decay & 0.0001 & 0.0001 & 0.0001 \\
		\hline
		\hline
		\textbf{Mixup} \\
		\hline
		Probability & 0.5  & 0.5  & 0.5 \\
		$\alpha$ & 1 & 1 & 1 \\
		\hline
		\hline
		\textbf{AutoAugment} \\
		\hline
		Probability & 0.5  & 0.5  & 0.5 \\
		Policy* & Cifar10 & Cifar10 & Imagenet \\
	\end{tabular}
	\caption{Selected hyperparameters for our experiments. "Evaluation" refers to the procedure described in Section \ref{sec:methodology}.
	(*)AutoAugment policies taken from the PyTorch library.}
\end{table}

\section{All Results}\label{app:remaining_results}
\subsection{Cifar10}
\begin{figure}[H]
	\centering
	\includegraphics[width=0.8\linewidth]{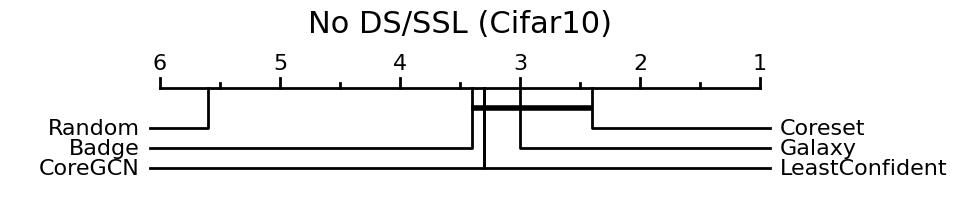}
	\caption{Ranking of AL methods on Cifar10 without DA or SSL}
\end{figure}
\begin{figure}[H]
	\centering
	\includegraphics[width=0.8\linewidth]{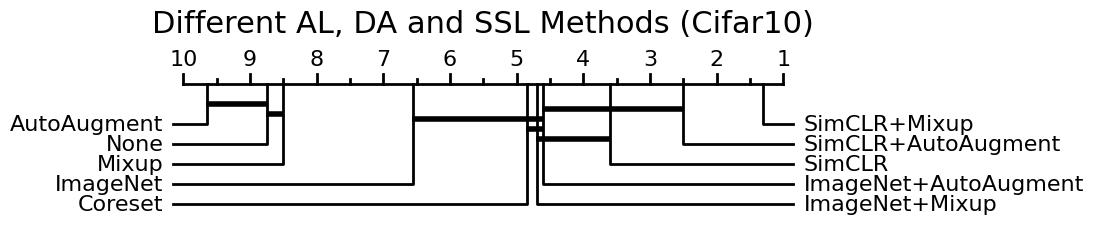}
	\caption{Ranking of \textbf{best} AL and different DA/SSL methods on Cifar10}
\end{figure}
\begin{figure}[H]
	\centering
	\includegraphics[width=\linewidth]{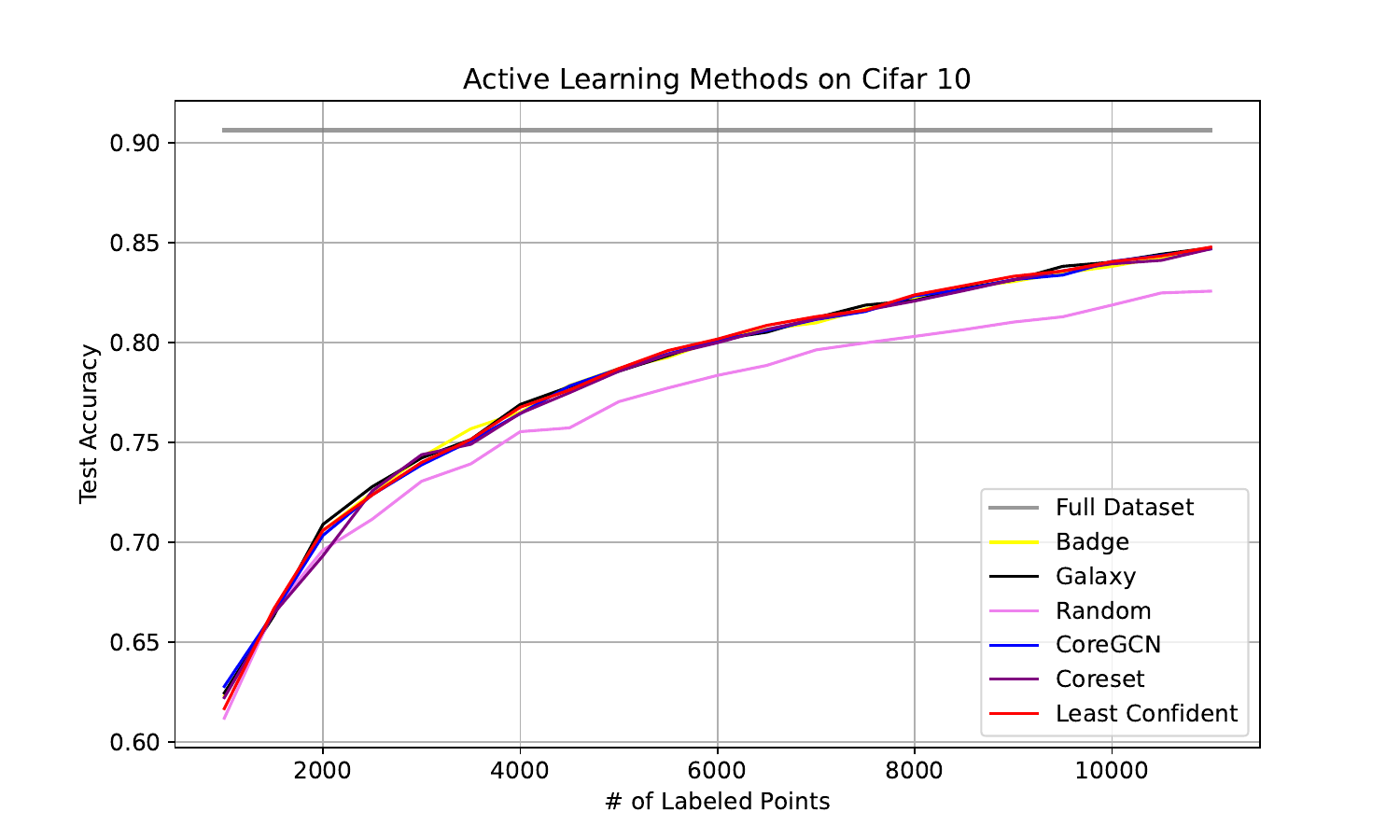}
	\caption{Test accuracy curves for Cifar10 with optimal combination of DA and SSL methods.}
\end{figure}
\begin{figure}[H]
	\centering
	\includegraphics[width=0.8\linewidth]{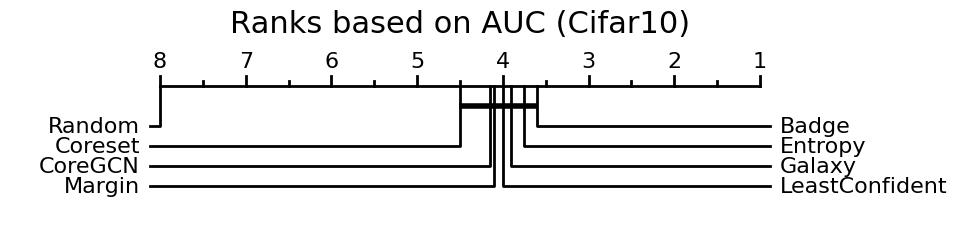}
	\caption{Ranking of AL methods on Cifar10 with optimal combination of DA and SSL methods.}
\end{figure}
\subsection{Cifar100}
\begin{figure}[H]
	\centering
	\includegraphics[width=0.8\linewidth]{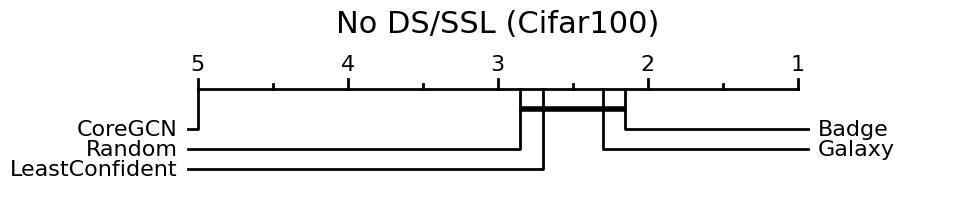}
	\caption{Ranking of AL methods on Cifar100 without DA or SSL}
\end{figure}
\begin{figure}[H]
	\centering
	\includegraphics[width=0.8\linewidth]{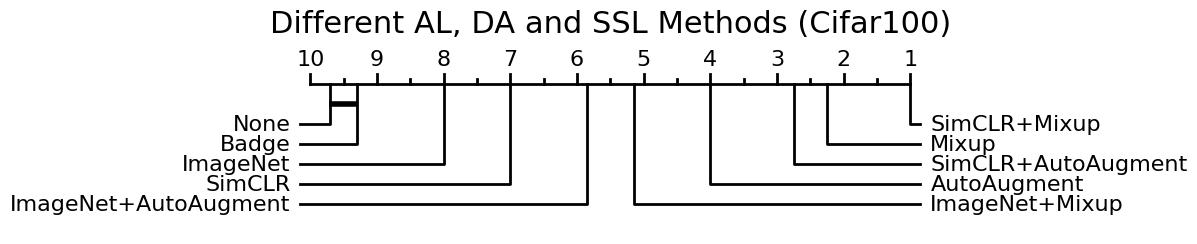}
	\caption{Ranking of \textbf{best} AL and different DA/SSL methods on Cifar100}
\end{figure}
\begin{figure}[H]
	\centering
	\includegraphics[width=\linewidth]{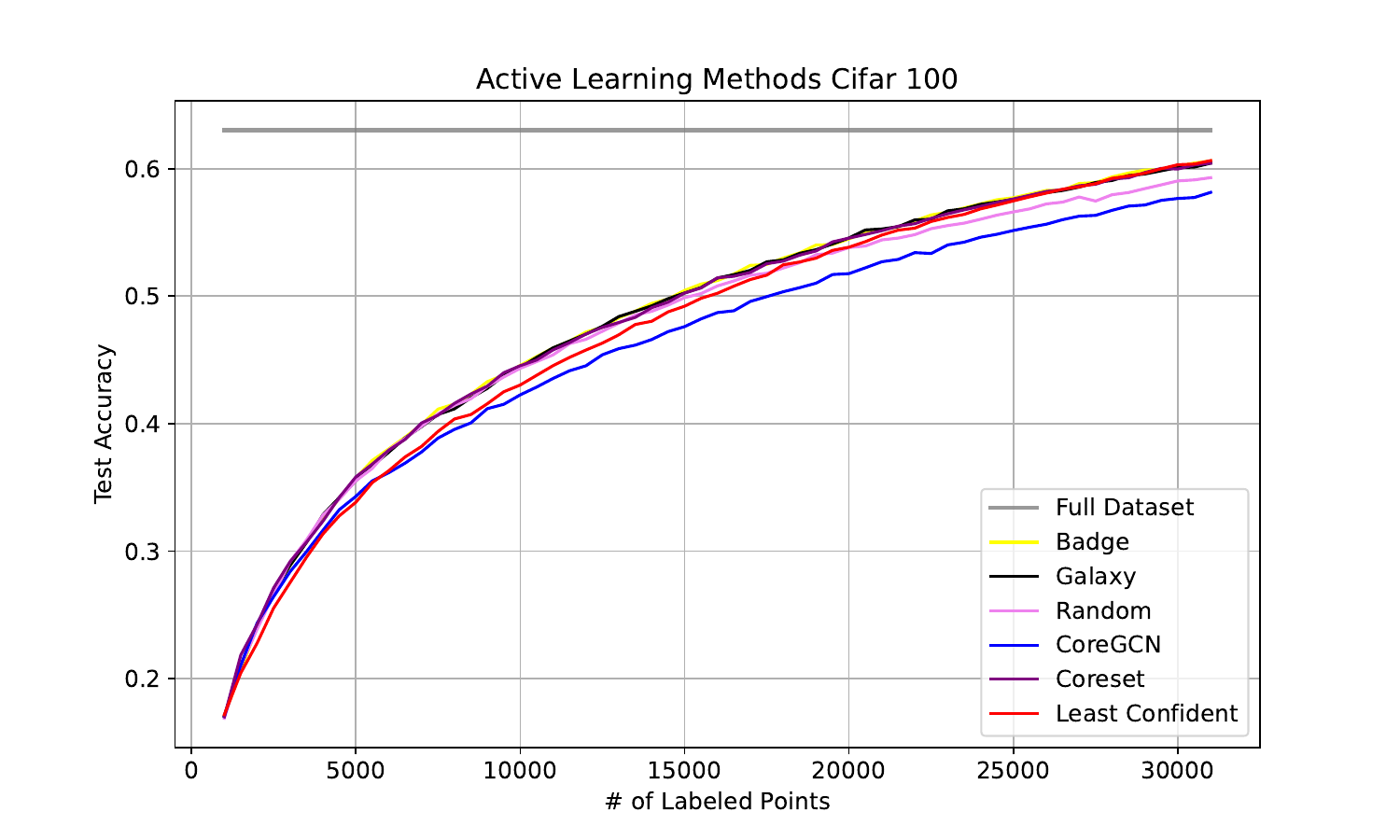}
	\caption{Test accuracy curves for Cifar100 with optimal combination of DA and SSL methods.}
\end{figure}
\begin{figure}[H]
	\centering
	\includegraphics[width=0.8\linewidth]{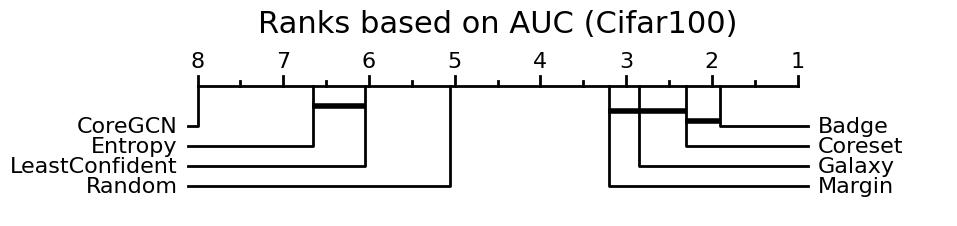}
	\caption{Ranking of AL methods on Cifar100 with optimal combination of DA and SSL methods.}
\end{figure}
\subsection{LSUN}
\begin{figure}[H]
	\centering
	\includegraphics[width=0.8\linewidth]{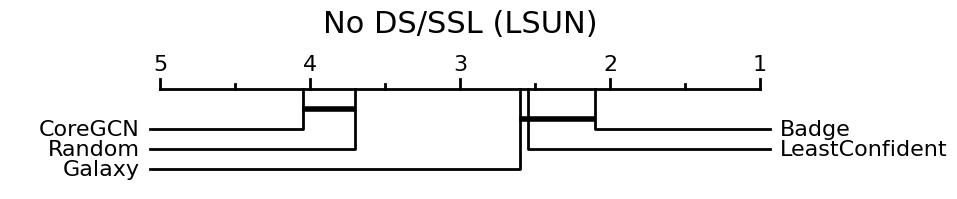}
	\caption{Ranking of AL methods on LSUN without DA or SSL}
\end{figure}
\begin{figure}[H]
	\centering
	\includegraphics[width=0.8\linewidth]{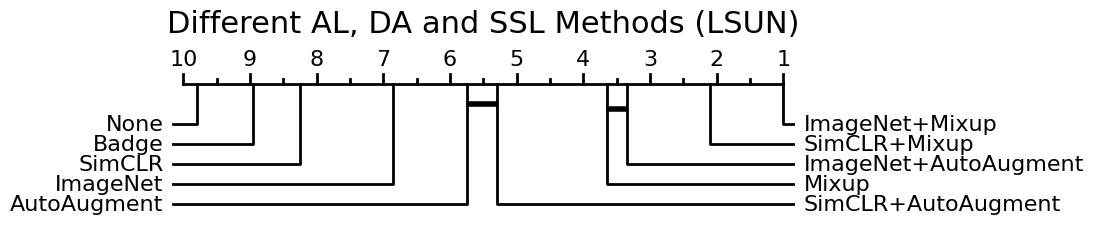}
	\caption{Ranking of \textbf{best} AL and different DA/SSL methods on LSUN}
\end{figure}
\begin{figure}[H]
	\centering
	\includegraphics[width=\linewidth]{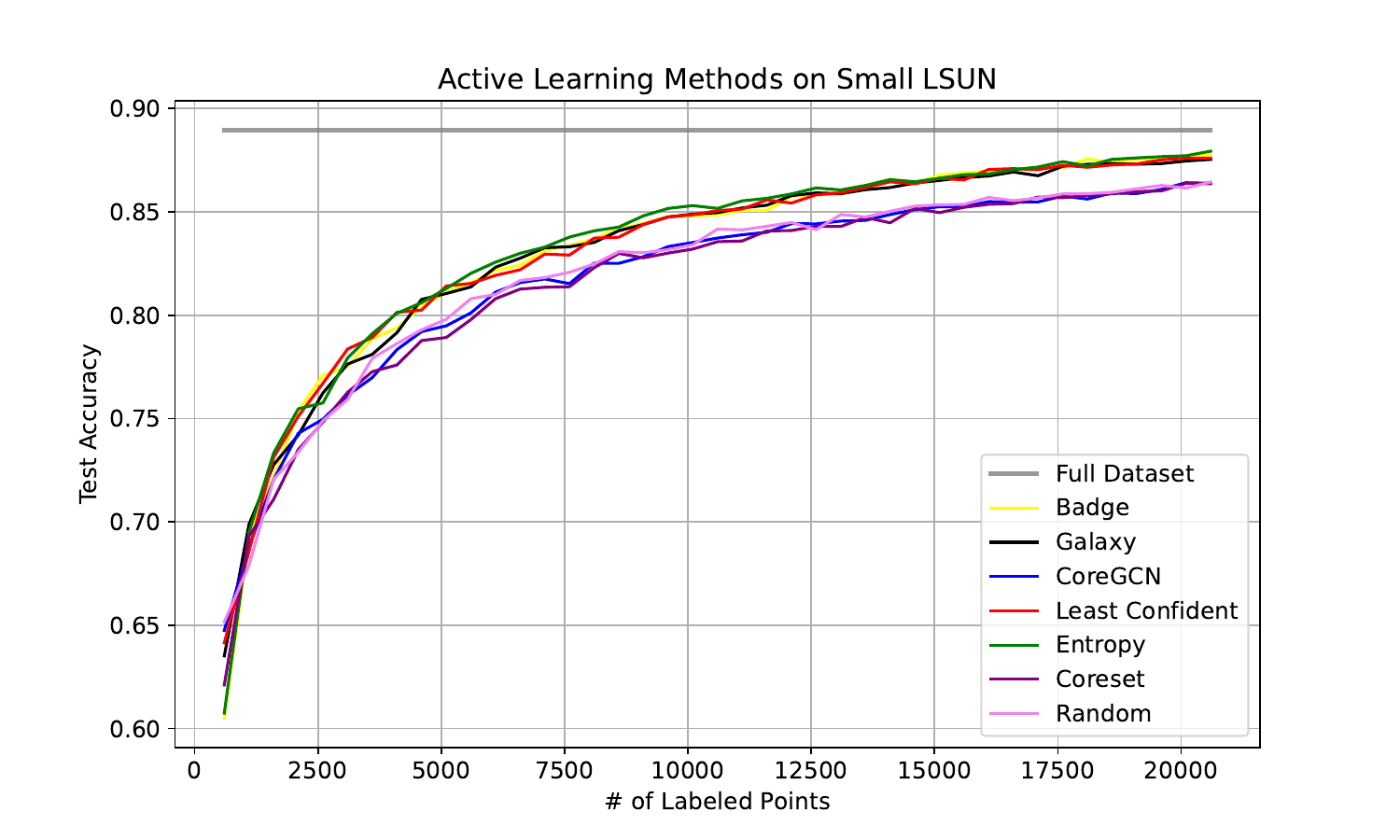}
	\caption{Test accuracy curves for LSUN with optimal combination of DA and SSL methods.}
\end{figure}
\begin{figure}[H]
	\centering
	\includegraphics[width=0.8\linewidth]{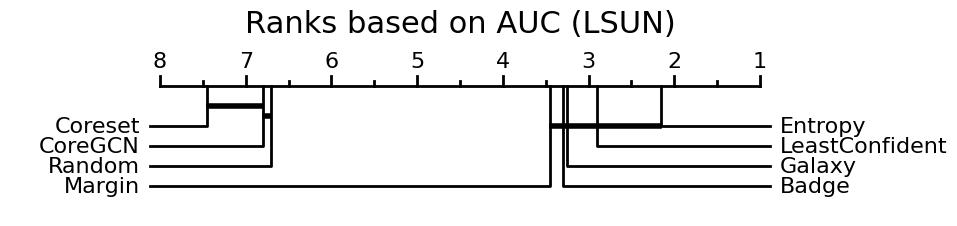}
	\caption{Ranking of AL methods on LSUN with optimal combination of DA and SSL methods.}
\end{figure}

\end{document}